\pdfoutput=1

\documentclass[11pt]{article}

\usepackage{acl}

\usepackage{times}
\usepackage{color}
\usepackage{latexsym}
\usepackage{amsmath}
\usepackage{booktabs}
\usepackage{multirow}
\usepackage{graphics}
\usepackage{graphicx} 
\usepackage{caption}
\usepackage{amssymb}
\usepackage{pifont}
\usepackage{subfigure}
\newcommand{\tabincell}[2]{\begin{tabular}{@{}#1@{}}#2\end{tabular}}
\usepackage[T1]{fontenc}

\usepackage[utf8]{inputenc}

\usepackage{microtype}
\usepackage{hyperref}
\title{LiLT: A Simple yet Effective Language-Independent Layout Transformer for Structured Document Understanding}

\author{Jiapeng Wang$^1$ \qquad  Lianwen Jin\thanks{\ \ Corresponding author.}\ $^{1,3,4}$  \qquad  Kai Ding$^{*2,3}$ \\
$^1$South China University of Technology, Guangzhou, China\\
$^2$IntSig Information Co.,  Ltd, Shanghai, China\\
$^3$INTSIG-SCUT Joint Laboratory of Document Recognition and Understanding, China\\
$^4$Peng Cheng Laboratory, Shenzhen, China\\
$^{1}$\texttt{eejpwang@mail.scut.edu.cn,  eelwjin@scut.edu.cn} \\
$^{2}$\texttt{danny\_ding@intsig.net}\\
}

\begin{document}
\maketitle
\begin{abstract}
Structured document understanding has attracted considerable attention and made significant progress recently, owing to its crucial role in intelligent document  processing. However, most existing related models can only  deal with the  document data of specific language(s)  (typically English) included in the pre-training collection,  which is extremely limited. To address this issue, we propose a simple yet effective \textbf{L}anguage-\textbf{i}ndependent \textbf{L}ayout \textbf{T}ransformer (\textbf{LiLT}) for structured 
document understanding. LiLT can be pre-trained on the structured documents of a single language and  then directly fine-tuned on other  languages with the corresponding off-the-shelf monolingual/multilingual pre-trained textual models. Experimental results on eight languages have shown that LiLT  can achieve  competitive or even superior performance on diverse widely-used downstream benchmarks, which enables language-independent benefit from the pre-training of document layout structure. Code and model 
are publicly available at  \href{https://github.com/jpWang/LiLT}{https://github.com/jpWang/LiLT}.
\end{abstract}

\section{Introduction}
Structured document understanding (SDU) aims at reading and analyzing the textual and structured information contained in scanned/digital-born documents. 
With the acceleration of the digitization process, it has been regarded as a crucial part of  intelligent document  processing and required by many real-world  applications in various industries such as finance, medical treatment and  insurance.

Recently, inspired by the rapid development of pre-trained language
models of plain texts \cite{devlin2019bert,liu2019roberta,bao2020unilmv2,chi2021infoxlm},
 many researches  on structured document  pre-training \cite{Layoutlm,layoutlmv2,xu2021layoutxlm,li2021structurallm,li2021selfdoc,li2021structext,appalaraju2021docformer}  have also pushed the limit of a variety of SDU tasks.
However, almost  all of them only focus on pre-training and fine-tuning on the documents in a single language, typically English. This is extremely limited for  other languages, especially in the case of lacking pre-training structured document data.

\begin{figure}[t!]
\centering
\subfigure[A form.]{
\begin{minipage}[]
{0.46\linewidth}
\centering
\includegraphics[width=1.7cm,height=2.4cm]{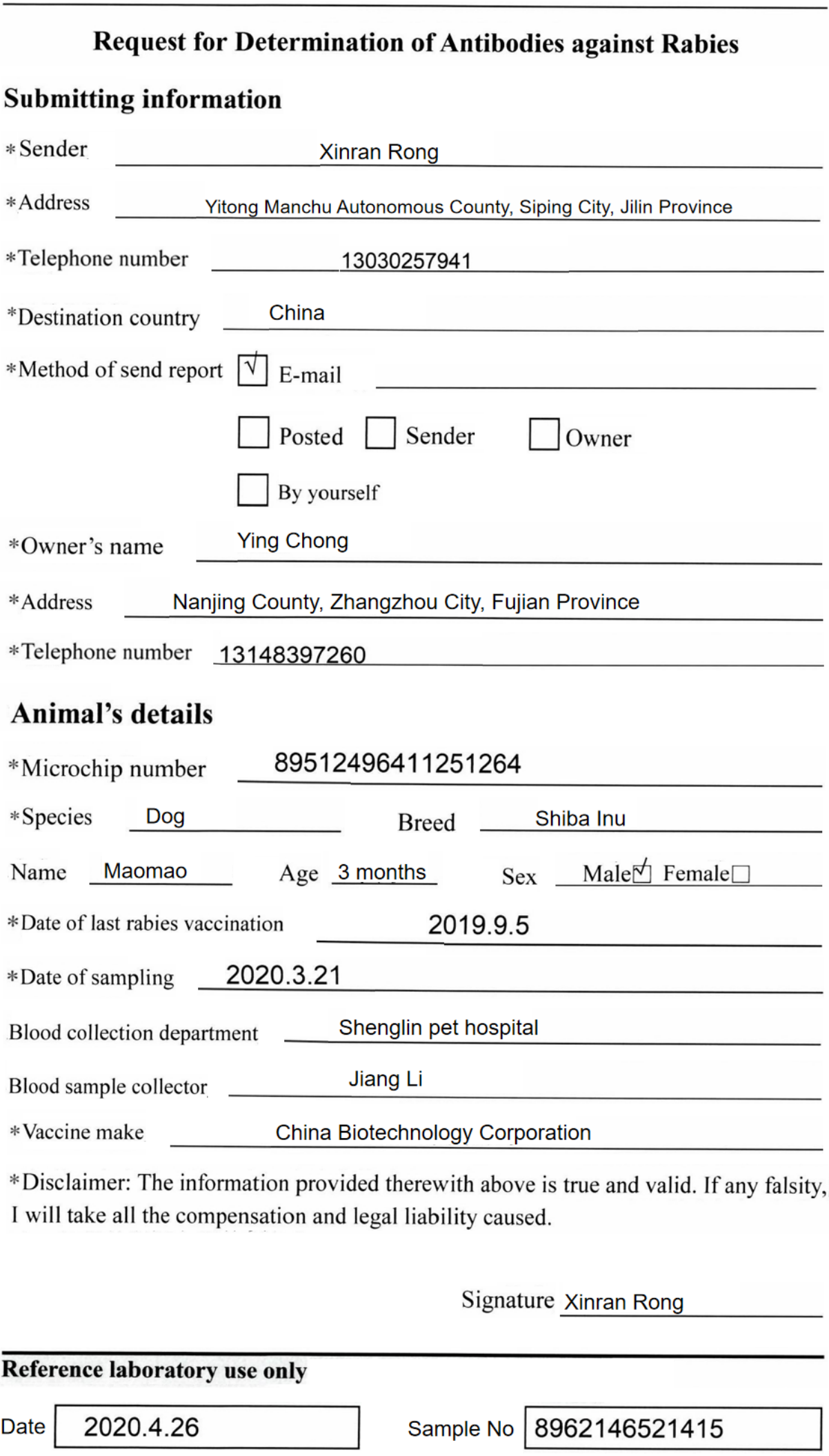}
\includegraphics[width=1.7cm,height=2.4cm]{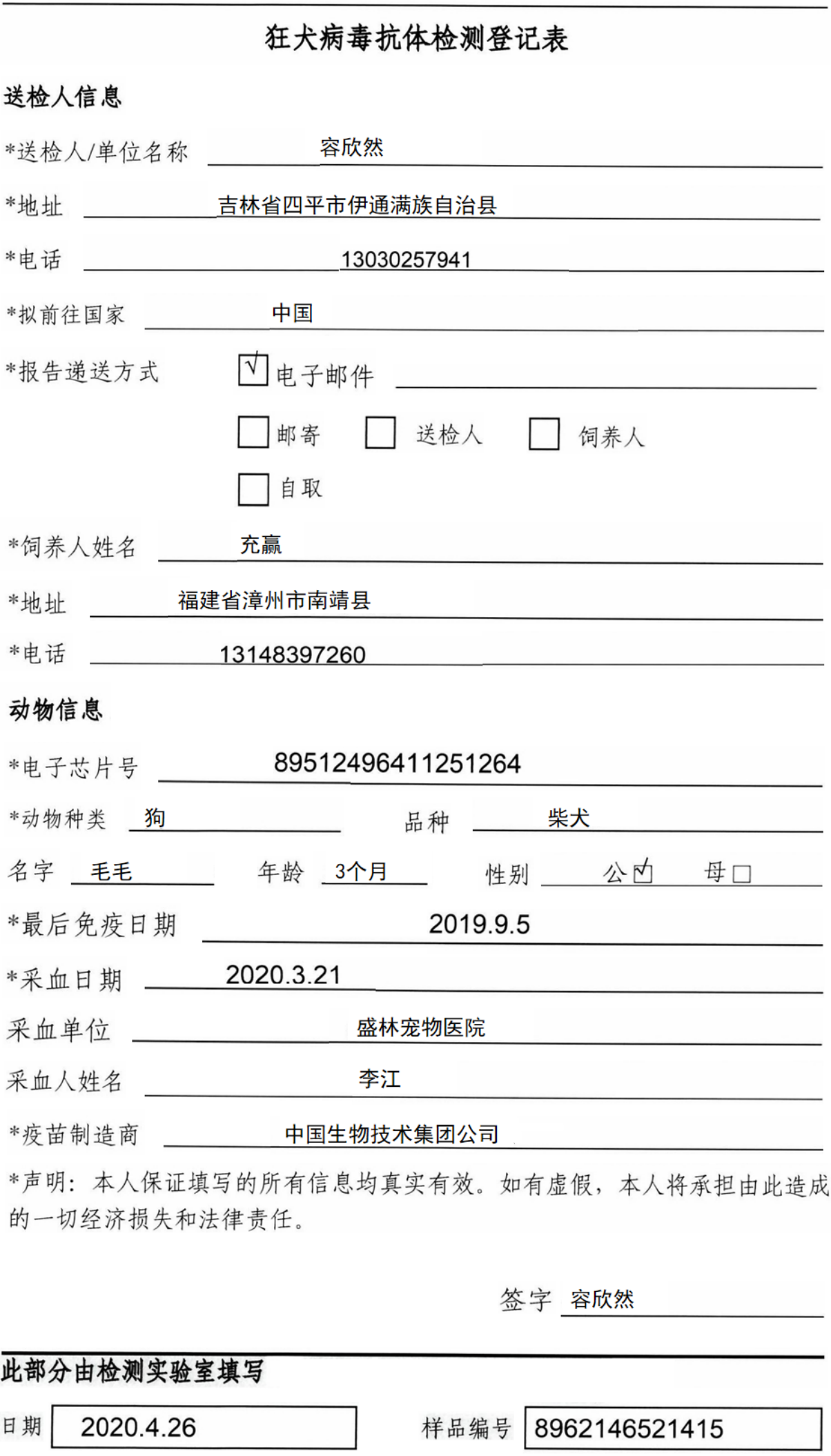}
\end{minipage}
}
\subfigure[A receipt.]{
\begin{minipage}[]
{0.46\linewidth}
\centering
\includegraphics[width=1.7cm,height=2.4cm]{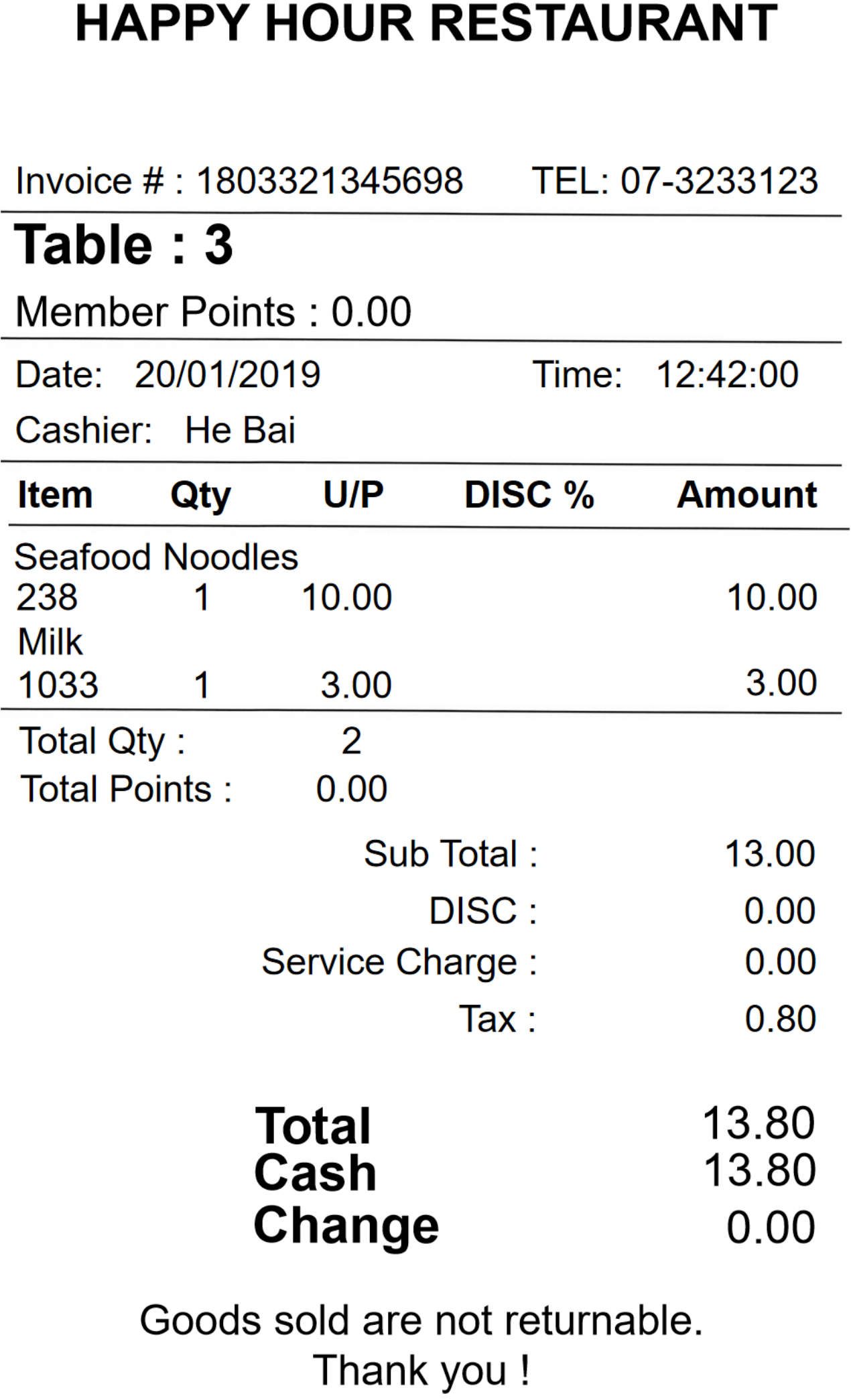}
\includegraphics[width=1.7cm,height=2.4cm]{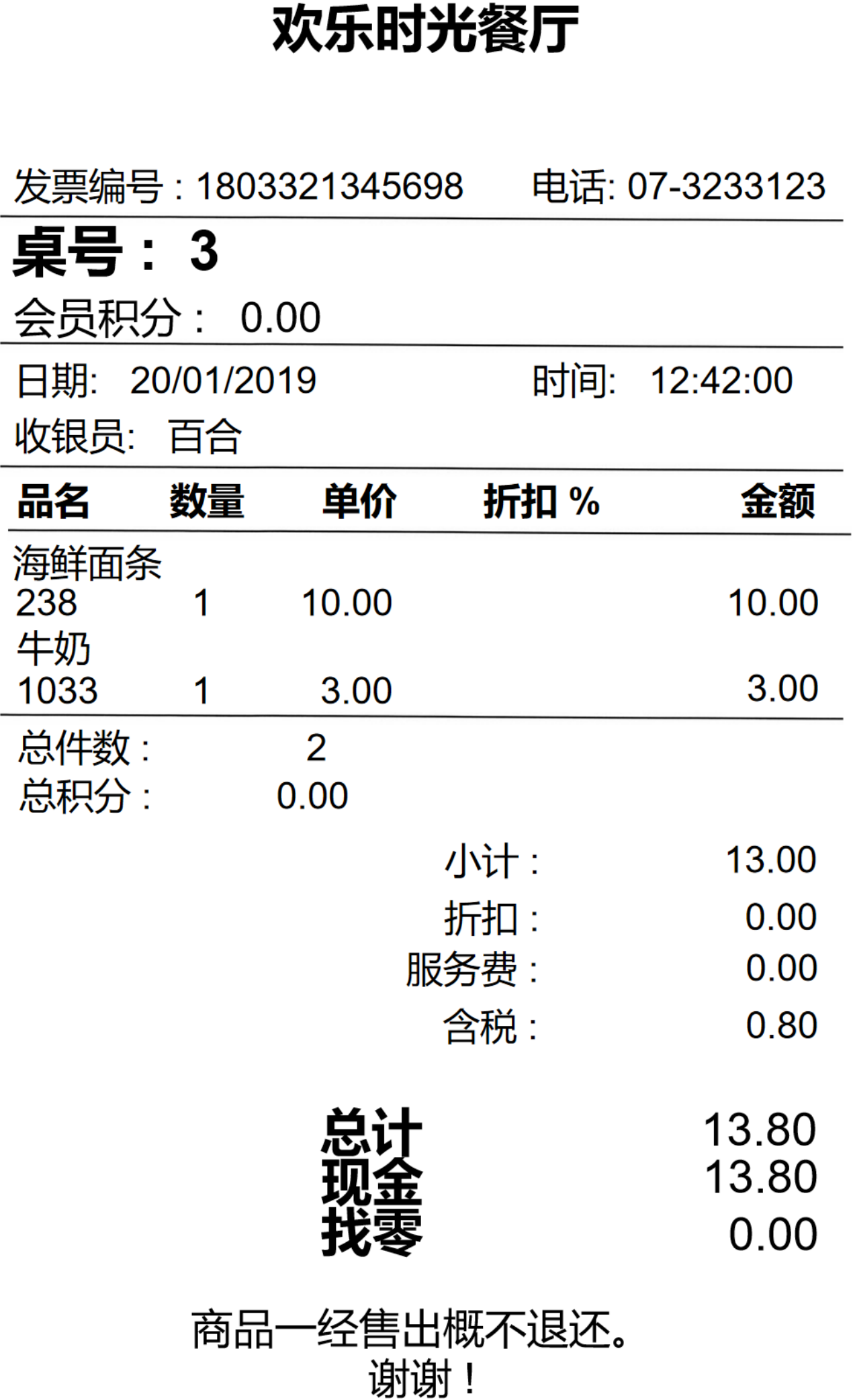}
\end{minipage}
}
\caption{The substitution of language does not appear obviously unnatural when the layout structure remains unchanged, as shown in a (a) form/(b) receipt. The detailed content has been re-synthesized to avoid the sensitive information leak. Best viewed in zoomed-in.}
\label{fig:docs}
\end{figure}

In this regard, we consider how to make the SDU tasks enjoy language-independent  benefit from   the   pre-training   of   document   layout structure. 
Here, we give an  observation as shown in Figure \ref{fig:docs}. When the layout structure remains unchanged, the substitution of language does not make obvious unnaturalness. It  fully motivates us to decouple and reuse the layout invariance among different languages.

Based on this inspiration, in this paper, we propose a simple yet effective \textbf{L}anguage-\textbf{i}ndependent \textbf{L}ayout \textbf{T}ransformer (\textbf{LiLT}) for structured document understanding.
In our framework, the  text and layout information are first decoupled and joint-optimized during pre-training, and then re-coupled for fine-tuning. 
To ensure that the two modalities have sufficient language-independent interaction, we further propose a novel  bi-directional attention complementation mechanism (BiACM) to enhance the cross-modality cooperation. 
Moreover, we present the key point location (KPL) and 
cross-modal alignment identification (CAI)
tasks, which are combined with the widely-used masked visual-language modeling (MVLM) to serve as our pre-training objectives. 
During fine-tuning, 
the layout flow (LiLT) can be  separated and  combined with  the off-the-shelf pre-trained textual models (such as 
RoBERTa \cite{liu2019roberta}, XLM-R \cite{conneau2020unsupervised}, InfoXLM \cite{chi2021infoxlm}, etc) to deal with the downstream tasks. In this way, our method decouples and learns the layout knowledge from the monolingual structured documents before generalizing  it  to the multilingual ones.

To the best of our knowledge,
the only pre-existing multilingual SDU model is LayoutXLM \cite{xu2021layoutxlm}. It scraps multilingual PDF documents of 53 languages from a web crawler and introduces extra pre-processing steps to clean the collected data, filter the low-quality  documents, and classify them into different languages.  After this, it utilizes a heuristic distribution to sample 22 million multilingual documents, which are further combined with the 8 million sampled English ones from the IIT-CDIP \cite{cdip} dataset (11 million English documents), resulting 30 million for pre-training with the LayoutLMv2 \cite{layoutlmv2} framework. However, this process is time-consuming and laborious.
On the contrary,  LiLT can be pre-trained with only IIT-CDIP  and then adapted to other languages. 
In this respect, 
LiLT is the first language-independent  method for structured document understanding.

Experimental results on eight languages have shown that LiLT  can achieve 
competitive or even superior performance on diverse widely-used downstream benchmarks, 
which substantially benefits numerous real-world SDU  applications. Our main contributions can be summarized as follows:
\begin{itemize}
\item We introduce a simple yet effective language-independent layout Transformer called LiLT for monolingual/multilingual structured document understanding. 
\item We  propose BiACM to provide  language-independent cross-modality  interaction,
along with an effective asynchronous optimization strategy for textual and non-textual flows in pre-training. 
Moreover, we present two new pre-training objectives, namely 
KPL and CAI.
\item LiLT achieves competitive or even superior performance on  various widely-used downstream benchmarks of different languages under 
different settings, which fully demonstrates its effectiveness.
\end{itemize}

\begin{figure*}[t]
\centering
\includegraphics[width=1.0\textwidth]{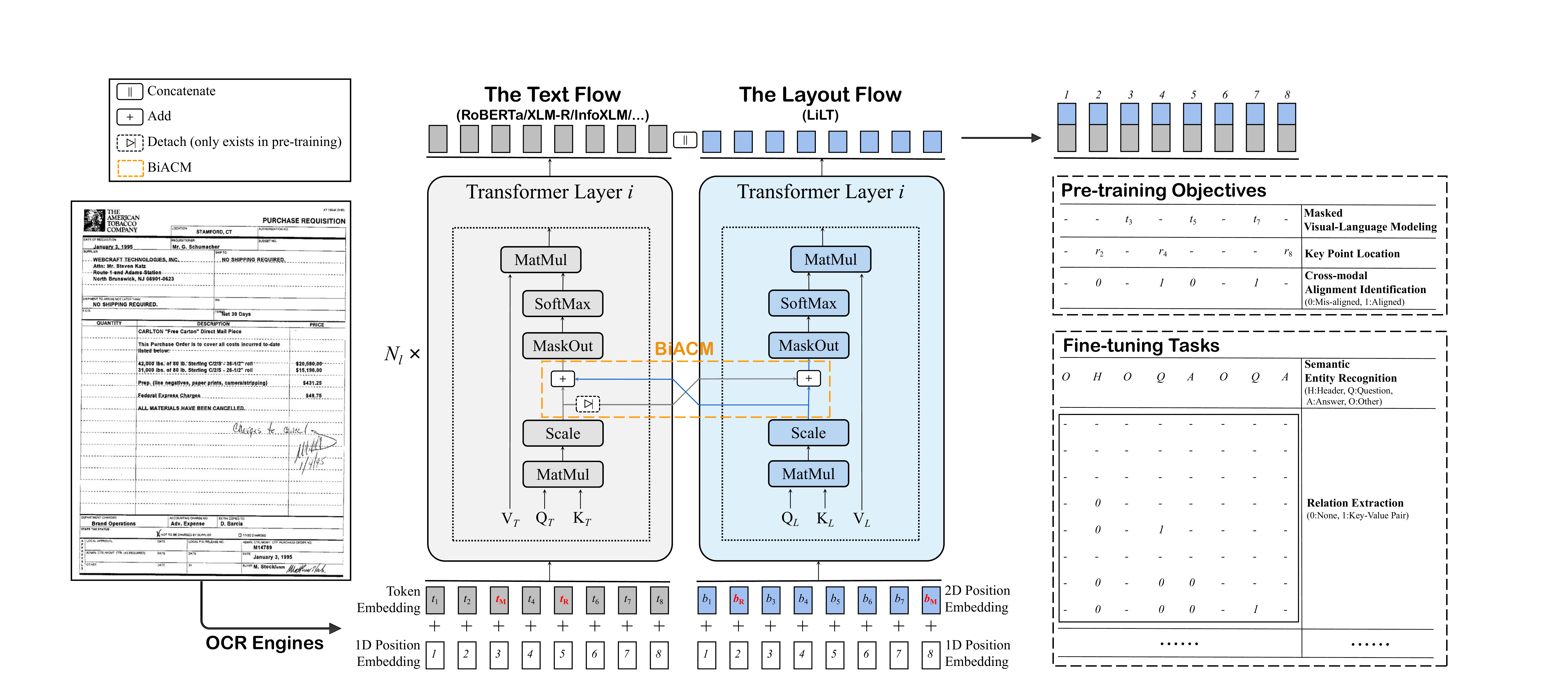}
\caption{The overall  illustration of our framework. 
Text and layout information are separately embedded and fed into the corresponding flow. 
BiACM is proposed to accomplish the cross-modality interaction. At the model output, text and layout features are concatenated for the self-supervised pre-training or the downstream fine-tuning.
$N_l$ is the number of Transformer layers.
The red *$_\mathrm{M}$/*$_\mathrm{R}$ indicates the randomly masked/replaced item for pre-training. $t$, $b$ and $r$ represent $token$, $box$ and $region$, respectively. Best viewed in zoomed-in.} 
\label{pipeline}
\end{figure*}

\section{LiLT}
Figure \ref{pipeline} shows the overall illustration of our method. Given an input  document image, we first use off-the-shelf OCR engines to get  text bounding boxes and  contents. Then, the text and layout information are separately embedded and fed into the corresponding Transformer-based architecture to obtain enhanced features. 
Bi-directional attention complementation mechanism (BiACM) is introduced to  accomplish the cross-modality interaction of text and layout clues.
Finally, the  encoded text and layout features are concatenated and additional heads are added upon them, for the self-supervised pre-training or the downstream fine-tuning.

\subsection{Model Architecture}
The whole framework can be regarded as a parallel dual-stream Transformer. The layout flow shares a similar structure as text flow, except for the reduced hidden size and intermediate size to achieve computational efficiency.

\subsubsection{Text  Embedding}
Following the common practice \cite{devlin2019bert,Layoutlm}, in the text flow, all text strings in the OCR results are first tokenized and concatenated as a sequence $S_t$ by sorting the corresponding text bounding boxes from the top-left to bottom-right. Intuitively, the special tokens \texttt{[CLS]} and \texttt{[SEP]} are also added at the beginning and end of the sequence respectively.  After this, $S_t$ will be truncated or padded with extra \texttt{[PAD]} tokens until its length equals  the maximum sequence length $N$. Finally, we sum the  token embedding $E_{token}$ of $S_t$ and the 1D positional embedding $P_{1\mathrm{D}}$ to obtain the text embedding $E_{T}\in\mathcal{R}^{N\times d_T}$  as:
\begin{align}
    E_{T} = \mathrm{LN}(E_{token} + P_{1\mathrm{D}}),
\end{align}where $d_T$ is the number of text feature dimension and $\mathrm{LN}$ is the layer normalization \cite{ln}.

\subsubsection{Layout  Embedding}
As for the layout flow, we construct a 2D position sequence $S_l$ with the same length as the token sequence $S_t$ using the corresponding text bounding boxes. To be specific,  we normalize and discretize all box coordinates to integers in  the range $[0, 1000]$, and use four embedding layers to generate $x$-axis, $y$-axis, height, and width features separately. Given the normalized bounding boxes $B = \mathrm{(}x_{min}, x_{max}, y_{min}, y_{max}, width, height\mathrm{)}$, the 2D positional embedding $P_{2\mathrm{D}}\in\mathcal{R}^{N\times d_L}$ (where $d_L$ is the number of layout feature dimension) is  constructed  as follows:
\begin{align}
P_{2\mathrm{D}} = Linear(\mathrm{CAT}(&E_{x_{min}}, E_{x_{max}},\nonumber\\ E_{y_{min}}, E_{y_{max}}, &E_{_{width}}, E_{_{height}})). 
\end{align}
Here, the $E$s are embedded vectors. $Linear$ is a linear projection layer and $\mathrm{CAT}$ is the channel-wise concatenation operation. The special tokens \texttt{[CLS]}, \texttt{[SEP]} and \texttt{[PAD]} are  also  attached with (0,0,0,0,0,0), (1000,1000,1000,1000,0,0) and (0,0,0,0,0,0) respectively.
It is worth mentioning that,  for each token, we directly utilize the bounding box of the text string it belongs to, because the fine-grained  token-level information is not always included in the results of some OCR engines.

Since Transformer layers are permutation-invariant, here we introduce the  1D positional embedding again. The resulting layout embedding $E_{L}\in\mathcal{R}^{N\times d_L}$ can be formulated as:
\begin{align}
    E_{L} = \mathrm{LN}(P_{2\mathrm{D}} + P_{1\mathrm{D}}).
\end{align}
\subsubsection{BiACM}
The text embedding $E_T$ and layout embedding $E_L$ are fed into their respective sub-models to generate high-level enhanced features. 
However, it will considerably ignore 
the cross-modal interaction process
if we simply combine the text and layout features at the encoder output only.
The network also needs to comprehensively analyse them at earlier stages. 
In view of this, we propose a new bi-directional attention complementation mechanism (BiACM) to  strengthen the cross-modality interaction across the entire encoding pipeline. Experiments in Section \ref{ablation} will further verify its effectiveness.

The vanilla self-attention mechanism in Transformer layers  captures the correlation between query $x_i$ and key $x_j$ by projecting the two vectors and calculating the attention score  as:
\begin{align}
    \alpha_{ij}=\frac{{(x_iW^Q)(x_jW^K)}^\top}{\sqrt{d^{h}}}.
\end{align} 
Here, the description is for a single head in a single self-attention layer with hidden size of $d^h$ and projection metrics $W^Q$, $W^K$ for simplicity. 
Given $\alpha^{T}_{ij}$ and $\alpha^{L}_{ij}$ of the text and layout flows located in the same head of the same layer, BiACM shares them as  common knowledge,  which is  formulated as:
\begin{align}
&\widetilde{\alpha^{T}_{ij}}=\alpha^{L}_{ij}+\alpha^{T}_{ij}, \\
&\widetilde{\alpha^{L}_{ij}}=
\begin{cases}
\alpha^{L}_{ij}+\mathrm{DETACH}(\alpha^{T}_{ij}) \ \ \ \normalsize{\text{ $if \ $ $\mathrm{Pre}$-$\mathrm{train}$},}\\
\alpha^{L}_{ij}+\alpha^{T}_{ij}  \ \ \ \ \ \ \ \ \ \ \ \ \ \ \ \ \ \ \ \ \ \ \ \normalsize{\text{ $if \ $ $\mathrm{Fine}$-$\mathrm{tune}$}.}
\end{cases}
\label{detach}
\end{align}
In order to maintain the ability of LiLT to cooperate with different off-the-shelf text models in fine-tuning  as much as possible, we heuristically adopt the detached $\alpha^{T}_{ij}$ for $\widetilde{\alpha^{L}_{ij}}$, so that the  textual stream  will not be affected by the gradient of  non-textual one during pre-training, and its overall consistency can be preserved. Finally, the modified attention scores are used to weight the projected value vectors for subsequent modules in both flows.

\subsection{Pre-training Tasks}
We conduct three self-supervised pre-training tasks to guide the model   to autonomously  learn  joint representations with  cross-modal cooperation. The details are introduced below.
\subsubsection{Masked Visual-Language Modeling}
This task is originally derived from  \cite{devlin2019bert}.
MVLM randomly masks some of the input tokens and the model is asked to recover them  over the whole vocabulary using the output encoded features, driven by a cross-entropy loss. 
Meanwhile, the non-textual information remains unchanged. MVLM improves model learning on the language side with cross-modality information. The given layout embedding can also help the model better capture both inter- and intra-sentence relationships. We mask 15\% text tokens, among which 80\% are replaced by the special token \texttt{[MASK]}, 10\% are replaced by  random tokens sampled from the whole vocabulary, and 10\% remain the same.

\subsubsection{Key Point Location}
We propose this  task to make the model better understand  layout information in the structured documents. KPL equally divides the entire layout into several regions (we set 7$\times$7=49 regions by default)  and randomly masks some of the input bounding boxes. The model is required to predict  which  regions the key points (top-left corner, bottom-right corner, and center point) of each box belong to using separate heads.  To deal with it, the model is required to fully understand the text content and know where to put a specific word/sentence when the surrounding ones are given.  We mask 15\% boxes, among which 80\% are replaced by (0,0,0,0,0,0), 10\% are replaced by random boxes sampled from the same batch, and 10\% remain the same. Cross-entropy loss is adopted.

Since there may exist detection errors in the output of OCR engines, we let the model predict the discretized regions (as mentioned above) instead of the exact location. This strategy can moderately relax the punishment criterion while improving the model performance.

\subsubsection{Cross-modal Alignment Identification}
We collect those encoded features of token-box pairs that are masked and further replaced (mis-aligned) or kept unchanged (aligned) by MVLM and KPL, 
and build  an additional head upon them to identify whether each pair is aligned. To achieve this, the model is required to learn the cross-modal perception capacity. CAI is a binary classification task, and  a cross-entropy loss is applied for it.

\subsection{Optimization Strategy}
Utilizing a unified learning rate for all model parameters to perform the end-to-end training process is the most common optimization strategy. While in our case, 
it will cause the layout flow to continuously update in the direction of coupling with the evolving text flow in the pre-training stage, which is harmful to the  ability of LiLT to cooperate with different off-the-shelf  textual models during fine-tuning. Based on this consideration, we explore multiple ratios to greatly slow down  the pre-training optimization of the text stream.
We also find that an appropriate  reduction ratio is better than  parameter freezing.

Note that, we adopt a unified learning rate for end-to-end optimization during fine-tuning. The DETACH operation of BiACM is also canceled at this time, as shown in Equation \ref{detach}.

\section{Experiments}
\subsection{Pre-training Setting}
We pre-train LiLT on the IIT-CDIP Test Collection 1.0 \cite{cdip}, which is a large-scale scanned document image dataset and contains more than 6 million documents with more than 11 million scanned document images. 
We  use TextIn API$\footnote{\href{https://www.textin.com}{https://www.textin.com}}$
to obtain the text bounding boxes and strings for this dataset. 
 
In this paper,
we initialize the text flow from the existing pre-trained English RoBERTa$\mathrm{_{BASE}}$ \cite{liu2019roberta} for our document pre-training,  and combine LiLT$\mathrm{_{BASE}}$ with the pre-trained  InfoXLM$\mathrm{_{BASE}}$ \cite{chi2021infoxlm}/a new pre-trained RoBERTa$\mathrm{_{BASE}}$ for multilingual/monolingual fine-tuning.
They have an equal number of self-attention layers, attention heads and maximum sequence length, 
which ensures that BiACM can work normally.  
In this $\mathrm{BASE}$ setting, LiLT  has a 12-layer encoder with 192 hidden size, 768 feed-forward filter size and 12 attention heads, resulting in the number of parameters as 6.1M. The maximum sequence length $N$ is set as 512.

LiLT$\mathrm{_{BASE}}$ is pre-trained  using Adam optimizer \cite{adam,adamw}, with the learning rate  2$\times$10$^{-5}$, weight decay  1$\times$10$^{-2}$, and ($\beta_1$, $\beta_2$) = (0.9, 0.999). The learning rate is linearly warmed up over the first 10\% steps and then linearly decayed. We set the  batch size as 96 and train LiLT$\mathrm{_{BASE}}$ for 5 epochs on the IIT-CDIP dataset using 4 NVIDIA A40 48GB GPUs.

\begin{table}[t]
\centering
\scalebox{0.78}{\begin{tabular}{clc}
\toprule
\textbf{\#} & \textbf{Inter-modal Operation} 
& \textbf{Average F1}\\
\midrule
\midrule
1  &   CAT &  0.6751   \\
2     & CAT$+$Co-Attention \cite{lu2019vilbert} &  0.6276   \\
3  & CAT$+$BiACM & \textbf{0.7963}    \\
4    & CAT$+$BiACM$-$DETACH in pre-training &  0.7682  \\
5    & CAT$+$BiACM$+$DETACH in fine-tuning &  0.7822 \\
\midrule
6 & \tabincell{l}{The text flow alone\\(InfoXLM$\mathrm{_{BASE}}$, as shown in Table \ref{tab:language-specific})}  & 0.7207\\
\bottomrule
\end{tabular}}
\\(a) BiACM. CAT is short for concatenation.
\bigskip 

\setlength{\tabcolsep}{5.7mm}{
\scalebox{0.72}{\begin{tabular}{cccccc}
\toprule
\textbf{\#} & \textbf{MVLM} & \textbf{KPL} & \textbf{CAI} & \textbf{Average F1} \\
\midrule
\midrule
1 & \ding{51}   &   &    &  0.7616   \\
2 & \ding{51}   & \ding{51}  &    &  0.7748   \\
3 & \ding{51}   &    & \ding{51}   &  0.7809   \\
4 & \ding{51}   & \ding{51}  & \ding{51}   &  \textbf{0.7963}   \\
\bottomrule
\end{tabular}}
}
\\(b) Pre-training tasks.
\bigskip 

\setlength{\tabcolsep}{7.0mm}{
\scalebox{0.72}{\begin{tabular}{clcc}
\toprule
\textbf{\#} & \textbf{Slow-down Ratio}   & \textbf{Average F1}  \\
\midrule
\midrule
1 & 1 (No Slow-down)  & 0.7840  \\
2 & 500  & 0.7901  \\
3 & 800  & 0.7947 \\
4 & 1000 & \textbf{0.7963}  \\
5 & 1200 & 0.7935  \\
6 & $+\infty$ (Parameter Freezing)  & 0.7893  \\
\bottomrule
\end{tabular}}
}
\\(c) Slow-down ratios.
\caption{
Ablation study of LiLT$\mathrm{_{BASE}}$ combined with InfoXLM$\mathrm{_{BASE}}$  \cite{chi2021infoxlm} on  the FUNSD and XFUND datasets (8 languages in total). The average F1 accuracy of language-specific semantic entity recognition (SER) task is given. (a) BiACM. (b) Pre-training tasks. (c) Slow-down ratios of the pre-training optimization for the text flow.}
\label{tab:ablation}
\end{table}

\begin{table}[!]
\centering
\scalebox{0.78}{
\begin{tabular}{lclcccc}
\toprule
 \textbf{Model}  &\textbf{Precision} & \textbf{Recall} & \textbf{F1} \\
\midrule
\midrule
BERT$\mathrm{_{BASE}}^1$ &0.5469 &0.6710   & 0.6026     \\
RoBERTa$\mathrm{_{BASE}}^2$   & 0.6349 &0.6975   & 0.6648    \\
UniLMv2$\mathrm{_{BASE}}^3$   &0.6349 &0.6975 &0.6648 \\
\midrule
LayoutLM$\mathrm{_{BASE}}^4$    &0.7597 &0.8155   & 0.7866 \\
BROS$\mathrm{_{BASE}}^5$    &0.8056 &0.8188   & 0.8121 \\
SelfDoc$^6$  &-&-  & 0.8336   \\
LayoutLMv2$\mathrm{_{BASE}}^7$    &0.8029 &0.8539 & 0.8276  \\
StrucTexT$\mathrm{_{BASE}}^8$    &\underline{0.8568} &0.8097 & 0.8309  \\
DocFormer$\mathrm{_{BASE}}^9$    &0.8076 &0.8609 & 0.8334  \\
\textbf{$^\star$}LayoutXLM$\mathrm{_{BASE}}^{10}$     &0.7913 &0.8158 & 0.8034  \\

\midrule
\textbf{LiLT[}EN-R$^2$\textbf{]}$\mathrm{_{BASE}}$   &\textbf{0.8721} &\textbf{0.8965}   & \textbf{0.8841}  \\
\textbf{$^\star$}\textbf{LiLT[}InfoXLM$^{11}$\textbf{]}$\mathrm{_{BASE}}$   &0.8467& \underline{0.8709}   & \underline{0.8586}  \\
\bottomrule
\end{tabular}}
\caption{Comparison on the semantic entity recognition (SER)  task of FUNSD \cite{funsd} dataset. \textbf{Bold}
indicates the SOTA and \underline{underline} indicates the second best. ``EN-R” is short for  English RoBERTa. \textbf{$^\star$}The multilingual model. \textbf{[]} denotes the off-the-shelf textual model used as the text flow of LiLT. $^1$\cite{devlin2019bert};$^2$\cite{liu2019roberta};$^3$\cite{bao2020unilmv2};$^4$\cite{Layoutlm};$^5$\cite{hong2020bros};$^6$\cite{li2021selfdoc};$^7$\cite{layoutlmv2};$^8$\cite{li2021structext};$^9$\cite{appalaraju2021docformer};$^{10}$\cite{xu2021layoutxlm};$^{11}$\cite{chi2021infoxlm}.}
\label{tab:funsd}
\end{table}

\subsection{Ablation Study}\label{ablation}
Considering  the complete pre-training takes a   long time, we pre-train LiLT$\mathrm{_{BASE}}$ with  2M documents randomly sampled from IIT-CDIP for 5 epochs to conduct ablation experiments, as shown in  
Table \ref{tab:ablation}.

We first evaluate the effect of introducing BiACM. In setting (a)\#1, the text and layout features are concatenated at the model output without any further interaction.
Compared with (a)\#6, we find that such a plain design results in a much worse performance than using the text flow alone. 
From  (a)\#1 to  (a)\#3, the significant improvement demonstrates that 
it is the novel BiACM that makes the transfer from ``monolingual” to ``multilingual” successful. Beside this, we have also tried to replace BiACM with the co-attention mechanism \cite{lu2019vilbert} which is widely adopted in  dual-stream Transformer architecture. It can be seen as a  ``deeper” cross-modal interaction, since  the keys and values from each modality are passed as input to the other modality’s dot-product attention calculation. However, severe drops are observed as shown in (a)\#2 vs (a)\#1\#3. We attribute it to  the damage of such a ``deeper”  interaction to the overall consistency of the text flow in the pre-training optimization. In contrast, BiACM can maintain LiLT's cross-model cooperation ability on the basis of providing cross-modal information.
Moreover, the necessity of DETACH in pre-training is  proved  in (a)\#4 vs (a)\#3.
Compared (a)\#3 to (a)\#5, we can also infer that removing DETACH in fine-tuning leads to a better performance.

\begin{table}[!]
\centering
\scalebox{0.80}{\begin{tabular}{lccc}
\toprule
 \textbf{Model} &\textbf{Precision} & \textbf{Recall} & \textbf{F1} \\
\midrule
\midrule
BERT$\mathrm{_{BASE}}$ &0.8833 &0.9107 &0.8968 \\
UniLMv2$\mathrm{_{BASE}}$  &0.8987 &0.9198 &0.9092 \\
\midrule
LayoutLM$\mathrm{_{BASE}}$  &0.9437 &0.9508 &0.9472 \\
BROS$\mathrm{_{BASE}}$  &0.9558 &0.9514 &0.9536 \\
LAMBERT$\mathrm{_{BASE}}^1$ &- &- &0.9441 \\
TILT$\mathrm{_{BASE}}^2$ &- &- &0.9511 \\
LayoutLMv2$\mathrm{_{BASE}}$  &0.9453 &0.9539 &0.9495 \\
DocFormer$\mathrm{_{BASE}}$    &\textbf{0.9652} & \underline{0.9614} & \textbf{0.9633}  \\
\textbf{$^\star$}LayoutXLM$\mathrm{_{BASE}}$  &0.9456 &0.9506 &0.9481 \\
\midrule
\textbf{LiLT[}EN-R\textbf{]}$\mathrm{_{BASE}}$   & \underline{0.9598} & \textbf{0.9616}   & \underline{0.9607} \\
\textbf{$^\star$}\textbf{LiLT[}InfoXLM\textbf{]}$\mathrm{_{BASE}}$   &  0.9574 & 0.9581  & 0.9577  \\
\bottomrule
\end{tabular}}
\caption{Comparison on the semantic entity recognition (SER)  task of CORD \cite{park2019cord} dataset. $^1$\cite{garncarek2020lambert};$^2$\cite{2021going}.}
\label{tab:cord}
\end{table}

Then, we compare the proposed KPL and CAI tasks. As shown in Table \ref{tab:ablation}(b), both tasks improve the model performance substantially, and the proposed CAI benefits the model more than KPL. 
Using both tasks together is more effective than using either one alone.

Finally, we explore the most suitable slow-down ratio for the pre-training optimization of the text flow. 
A ratio equal to 1 in (c)\#1 means there is no slow-down and a unified learning rate is adopted.
It can be found that the F1 scores  keep rising with the growth of slow-down ratios and begin to fall when the ratio is greater than 1000. Consequently, we set the slow-down ratio as 1000 by default.

\begin{table}[!]
\centering
\scalebox{0.81}{\begin{tabular}{lccc}
\toprule
 \textbf{Model} &\textbf{Precision} & \textbf{Recall} & \textbf{F1} \\
\midrule
\midrule

BiLSTM+CRF$^1$ &- &- & 0.8910 \\
GraphIE$^2$  &- &-& 0.9026 \\
GCN-based$^3$  &- &-& 0.9255 \\
TRIE$^4$  &- &-& 0.9321 \\
VIES$^5$  &- &-& 0.9523 \\
MatchVIE$^6$  &- &-& 0.9687 \\
TCPN$^7$  &- &-& 0.9759 \\
\midrule
RoBERTa$\mathrm{_{BASE}}^8$  &0.9405 &0.9640 &0.9521 \\
\midrule
StrucTexT$\mathrm{_{BASE}}$  &- &- & \underline{0.9795}  \\
\textbf{$^\star$}LayoutXLM$\mathrm{_{BASE}}$   &\underline{0.9699} & \underline{0.9820} & 0.9759 \\
\midrule
\textbf{LiLT[}ZH-R$^8$\textbf{]}$\mathrm{_{BASE}}$   &\textbf{0.9762} &\textbf{0.9833}   & \textbf{0.9797}  \\
\textbf{$^\star$}\textbf{LiLT[}InfoXLM\textbf{]}$\mathrm{_{BASE}}$   & \underline{0.9699} & \underline{0.9820}   & 0.9759  \\
\bottomrule
\end{tabular}}
\caption{Comparison on the semantic entity recognition (SER)  task of EPHOIE \cite{vies} dataset. ``ZH-R”  is short for  Chinese RoBERTa. $^1$\cite{blstmcrf};$^2$\cite{qian2019graphie};$^3$\cite{gcn};$^4$\cite{TRIE};$^5$\cite{vies};$^6$\cite{tang2021matchvie};$^7$\cite{tcpn};$^8$\cite{cui-etal-2020-revisiting}.}
\label{tab:ephoie}
\end{table}

\begin{table}[!t]
\centering
\setlength{\tabcolsep}{6,85mm}{
\scalebox{0.8}{\begin{tabular}{lcc}
\toprule
 \textbf{Model} & \textbf{Accuracy} \\
\midrule
\midrule
VGG-16$^1$  &90.97\% \\
Stacked CNN Single$^2$ &91.11\% \\
Stacked CNN Ensemble$^2$& 92.21\% \\
InceptionResNetV2$^3$ &92.63\% \\
LadderNet$^4$ &92.77\% \\
Multimodal Single$^5$& 93.03\% \\
Multimodal Ensemble$^5$ &93.07\% \\
\midrule
BERT$\mathrm{_{BASE}}$   &89.81\%   \\
UniLMv2$\mathrm{_{BASE}}$   &90.06\%  \\
\midrule
LayoutLM$\mathrm{_{BASE}}$ (w/ image)   &94.42\% \\
BROS$\mathrm{_{BASE}}$   &95.58\% \\
SelfDoc&93.81\% \\
TILT$\mathrm{_{BASE}}$  &93.50\% \\
LayoutLMv2$\mathrm{_{BASE}}$  &95.25\%  \\
DocFormer$\mathrm{_{BASE}}$    & \textbf{96.17\%}  \\
\textbf{$^\star$}LayoutXLM$\mathrm{_{BASE}}$    &95.21\% \\
\midrule
\textbf{LiLT[}EN-R\textbf{]}$\mathrm{_{BASE}}$   & \underline{95.68\%}  \\
\textbf{$^\star$}\textbf{LiLT[}InfoXLM\textbf{]}$\mathrm{_{BASE}}$  &95.62\%  \\
\bottomrule
\end{tabular}}
}
\caption{Comparison on the document classification (DC) task of RVL-CDIP \cite{rvl} dataset. $^1$\cite{rvl-vgg};$^2$\cite{das2018document};$^3$\cite{szegedy2017inception};$^4$\cite{ladder};$^5$\cite{ensemble}.}
\label{tab:rvl}
\end{table}

\begin{table*}[!]
\centering
\scalebox{0.70}{
\begin{tabular}{@{}clccccccccccc@{}}
\toprule
\multirow{2}{*}{Task} & \multicolumn{1}{c}{\multirow{2}{*}{Model}} & \multicolumn{2}{c}{Pre-training Docs} & FUNSD & \multicolumn{7}{c}{XFUND}         & \multirow{2}{*}{Avg.} \\ \cmidrule(l){3-4} \cmidrule(l){5-5} \cmidrule(l){6-12} &                        & Language           & Size           & EN    & ZH & JA & ES & FR & IT & DE & PT &                      \\ \midrule \midrule
\multirow{4}{*}{SER}  & XLM-RoBERTa$_{\mathrm{BASE}}$        &     -     &  -             &      0.6670& 0.8774 &0.7761& 0.6105& 0.6743& 0.6687 &0.6814 &0.6818 &0.7047 \\
& InfoXLM$_{\mathrm{BASE}}$            &     -    &   -            & 0.6852  & 0.8868  & 0.7865 & 0.6230 &  0.7015 &  0.6751  & 0.7063  & 0.7008  & 0.7207\\
& LayoutXLM$_{\mathrm{BASE}}$         &    Multilingual     &    30M           &  0.7940  & 0.8924  & 0.7921 &  0.7550  & 0.7902  & 0.8082 &  0.8222  & 0.7903  & 0.8056  \\
\cmidrule(l){2-13} 
& \textbf{LiLT[}InfoXLM\textbf{]}$_{\mathrm{BASE}}$             &     \textbf{English only}    &   \textbf{11M}          &   \textbf{0.8415}                     &    \textbf{0.8938}                 &   \textbf{0.7964}                  &   \textbf{0.7911}                  &   \textbf{0.7953}                  &       \textbf{0.8376}              &      \textbf{0.8231}               &       \textbf{0.8220}              &      \textbf{0.8251}                \\ \midrule  \midrule
\multirow{4}{*}{RE}  & XLM-RoBERTa$_{\mathrm{BASE}}$        &    -     &  -        &  0.2659 &  0.5105  & 0.5800  & 0.5295 &  0.4965  & 0.5305 &  0.5041 &  0.3982  & 0.4769  \\
& InfoXLM$_{\mathrm{BASE}}$            &    -   &   -       &  0.2920 &  0.5214  & 0.6000  & 0.5516 &  0.4913  & 0.5281 &  0.5262  & 0.4170  & 0.4910    \\
& LayoutXLM$_{\mathrm{BASE}}$         &    Multilingual     &    30M           &  0.5483  & 0.7073  & 0.6963 &  0.6896 &  0.6353  & 0.6415 &  0.6551 &  0.5718  & 0.6432 \\
\cmidrule(l){2-13} 
& \textbf{LiLT[}InfoXLM\textbf{]}$_{\mathrm{BASE}}$            &     \textbf{English only}    &   \textbf{11M}          &  \textbf{0.6276 }                     &   \textbf{0.7297   }               &     \textbf{0.7037   }             &        \textbf{0.7195 }            &    \textbf{0.6965     }            &          \textbf{0.7043  }         &         \textbf{0.6558   }         &      \textbf{0.5874    }           &  \textbf{0.6781}      \\ \bottomrule 
\end{tabular}}
\caption{Language-specific fine-tuning F1 accuracy on FUNSD and XFUND (fine-tuning on X, testing on X). “SER” denotes the semantic entity recognition and “RE” denotes the relation extraction.  \textbf{[]} indicates the off-the-shelf textual model used as the text flow of LiLT.}
\label{tab:language-specific}
\end{table*}

\subsection{Comparisons with the SOTAs}

To demonstrate the performance of LiLT, we conduct experiments on several widely-used monolingual datasets and the multilingual XFUND benchmark \cite{xu2021layoutxlm}. 
In addition to the experiments involving typical language-specific fine-tuning, 
we also follow the two settings designed in \cite{xu2021layoutxlm} to demonstrate the ability to transfer knowledge among different languages, which are zero-shot transfer learning and multitask fine-tuning, for fair comparisons. Specifically, (1) language-specific fine-tuning refers to the typical fine-tuning paradigm of fine-tuning on language X and testing on language X. (2) Zero-shot transfer learning means the models are fine-tuned on English data only and then evaluated on each target language. (3) Multitask fine-tuning requires the model to fine-tune on data in all languages. 
\subsubsection{Language-specific Fine-tuning}
We first evaluate LiLT on four  widely-used monolingual   datasets - FUNSD \cite{funsd}, CORD \cite{park2019cord}, EPHOIE \cite{vies} and  RVL-CDIP \cite{cdip}, and the results are shown in Table \ref{tab:funsd}, \ref{tab:cord}, \ref{tab:ephoie} and \ref{tab:rvl}.
We have found that (1) LiLT is flexible since it can work with monolingual or multilingual plain text models to deal with downstream tasks. (2) Although LiLT is designed for the transfer from ``monolingual” to ``multilingual”, it can surprisingly cooperate with 
monolingual textual models to achieve competitive or even superior performance (especially on the FUNSD dataset with only a few training samples available), compared with existing language-specific SDU  models such as LayoutLMv2 and DocFormer. (3) On these  datasets which are widely adopted for monolingual evaluation,  LiLT generally performs better than LayoutXLM.
This fully demonstrates the effectiveness of our pre-training framework and indicates that the layout and text information can  be  successfully  decoupled in pre-training and  re-coupled in fine-tuning.

Then we evaluate  LiLT on language-specific fine-tuning tasks of FUNSD and the multilingual XFUND \cite{xu2021layoutxlm}, and the results are shown in Table \ref{tab:language-specific}. Compared with the plain text models (XLM-R/InfoXLM) or the LayoutXLM model pre-trained with 30M multilingual structured documents, LiLT achieves the highest F1 scores on  both the SER and RE tasks of each language while using 11M monolingual data. This significant improvement shows LiLT’s capability to transfer language-independent knowledge from pre-training to downstream tasks.

\subsubsection{Zero-shot Transfer Learning}
The results of cross-lingual zero-shot transfer  are presented in  Table \ref{tab:zero-shot}. 
It can be observed that  the LiLT model transfers the most knowledge from English to other languages, and significantly outperforms its competitors. This fully verifies that LiLT can capture the common layout invariance among different languages.
Moreover, LiLT has never seen non-English  documents before evaluation under this setting, while the LayoutXLM model has been pre-trained with them. This is to say, LiLT faces a  stricter  cross-lingual zero-shot transfer scenario but achieves better performance. 
 
\begin{table*}[!]
\centering
\scalebox{0.68}{
\begin{tabular}{@{}clccccccccccc@{}}
\toprule
\multirow{2}{*}{Task} & \multicolumn{1}{c}{\multirow{2}{*}{Model}} & \multicolumn{2}{c}{Pre-training Docs} & FUNSD & \multicolumn{7}{c}{XFUND}         & \multirow{2}{*}{Avg.} \\ \cmidrule(l){3-4} \cmidrule(l){5-5} \cmidrule(l){6-12} &                        & Language           & Size           & EN    & ZH & JA & ES & FR & IT & DE & PT &                      \\ \midrule \midrule
\multirow{4}{*}{SER}  & XLM-RoBERTa$_{\mathrm{BASE}}$        &     -     &  -             &  0.6670& 0.4144 &0.3023& 0.3055 &0.3710& 0.2767& 0.3286 &0.3936& 0.3824  \\
& InfoXLM$_{\mathrm{BASE}}$            &     -    &   -            &  0.6852& 0.4408& 0.3603& 0.3102 &0.4021 &0.2880& 0.3587& 0.4502& 0.4119  \\
& LayoutXLM$_{\mathrm{BASE}}$         &    Multilingual     &    30M           & 0.7940 &0.6019& 0.4715& 0.4565& 0.5757& 0.4846& 0.5252& 0.5390& 0.5561   \\
\cmidrule(l){2-13} 
& \textbf{LiLT[}InfoXLM\textbf{]}$_{\mathrm{BASE}}\spadesuit$             &     \textbf{English only}    &   \textbf{11M}            &    \textbf{0.8415   }                 &    \textbf{0.6152    }             &   \textbf{ 0.5184   }              &    \textbf{ 0.5101    }            & \textbf{0.5923     }               &   \textbf{ 0.5371     }            &     \textbf{  0.6013         }     &   \textbf{  0.6325    }            &     \textbf{  0.6061   }                  \\ \midrule  \midrule
\multirow{4}{*}{RE}  & XLM-RoBERTa$_{\mathrm{BASE}}$        &     -     &  -             &  0.2659 &0.1601& 0.2611 &0.2440 &0.2240 &0.2374 &0.2288 &0.1996& 0.2276 \\
& InfoXLM$_{\mathrm{BASE}}$            &     -    &   -            &  0.2920& 0.2405 &0.2851 &0.2481 &0.2454& 0.2193 &0.2027 &0.2049 &0.2423  \\
& LayoutXLM$_{\mathrm{BASE}}$         &    Multilingual     &    30M           &  0.5483& 0.4494& 0.4408 &0.4708 &0.4416 &0.4090 &0.3820& 0.3685 &0.4388  \\
\cmidrule(l){2-13} 
& \textbf{LiLT[}InfoXLM\textbf{]}$_{\mathrm{BASE}}\spadesuit$             &     \textbf{English only}    &   \textbf{11M}           &    \textbf{0.6276}                   &     \textbf{0.4764}               &    \textbf{0.5081}                 &     \textbf{0.4968}            & \textbf{0.5209}               &   \textbf{0.4697}             &   \textbf{0.4169  }              &  \textbf{0.4272 }         &  \textbf{0.4930 }  \\ \bottomrule 
\end{tabular}}
\caption{Cross-lingual zero-shot transfer F1 accuracy  on FUNSD and XFUND (fine-tuning on FUNSD, testing on X). 
$\spadesuit$ indicates that LiLT faces a  stricter zero-shot transfer scenario compared with LayoutXLM, since it has never seen non-English documents before evaluation, even during pre-training.}
\label{tab:zero-shot}
\end{table*}
\subsubsection{Multi-task Fine-tuning}

Table \ref{tab:multitask} shows  the results of multitask learning. In this setting, the pre-trained LiLT model is simultaneously fine-tuned with all eight languages  and evaluated for each specific language.
We observe that this setting further improves the model performance compared to the language-specific fine-tuning, which confirms that SDU  can benefit from commonalities in the layout  of multilingual structured documents.
In addition, LiLT once again outperforms its counterparts by a large margin.

\begin{table*}[!]
\centering
\scalebox{0.68}{
\begin{tabular}{@{}clccccccccccc@{}}
\toprule
\multirow{2}{*}{Task} & \multicolumn{1}{c}{\multirow{2}{*}{Model}} & \multicolumn{2}{c}{Pre-training Docs} & FUNSD & \multicolumn{7}{c}{XFUND}         & \multirow{2}{*}{Avg.} \\ \cmidrule(l){3-4} \cmidrule(l){5-5} \cmidrule(l){6-12} &                        & Language           & Size           & EN    & ZH & JA & ES & FR & IT & DE & PT &                      \\ \midrule \midrule
\multirow{4}{*}{SER}  & XLM-RoBERTa$_{\mathrm{BASE}}$        &     -     &  -             &        0.6633 &  0.8830 &  0.7786 &  0.6223&   0.7035&   0.6814 &  0.7146&   0.6726 &  0.7149   \\
& InfoXLM$_{\mathrm{BASE}}$            &     -    &   -            &          0.6538&  0.8741&  0.7855&  0.5979&  0.7057 & 0.6826&  0.7055 & 0.6796&  0.7106      \\
& LayoutXLM$_{\mathrm{BASE}}$         &    Multilingual     &    30M           &     0.7924 &   0.8973  &  0.7964 &   0.7798 &   0.8173  &  0.8210 &   0.8322  &  0.8241  &  0.8201        \\
\cmidrule(l){2-13} 
& \textbf{LiLT[}InfoXLM\textbf{]}$_{\mathrm{BASE}}$              &     \textbf{English only}    &   \textbf{11M}        &  \textbf{0.8574     }                 &    \textbf{ 0.9047   }             &       \textbf{ 0.8088 }            &  \textbf{    0.8340    }           &  \textbf{ 0.8577   }               &  \textbf{ 0.8792   }               &  \textbf{     0.8769    }          &  \textbf{ 0.8493   }               &         \textbf{ 0.8585   }              \\ \midrule  \midrule
\multirow{4}{*}{RE}  & XLM-RoBERTa$_{\mathrm{BASE}}$        &     -     &  -             &   0.3638 &  0.6797  & 0.6829 &  0.6828 &  0.6727  & 0.6937 &  0.6887  & 0.6082  & 0.6341                     \\
& InfoXLM$_{\mathrm{BASE}}$            &     -    &   -            &    0.3699 &  0.6493  & 0.6473 &  0.6828 &  0.6831  & 0.6690 &  0.6384 &  0.5763 &  0.6145        \\
& LayoutXLM$_{\mathrm{BASE}}$         &    Multilingual     &    30M           &    0.6671 &  0.8241 &  0.8142 &  0.8104 &  0.8221  & 0.8310 &  0.7854  & 0.7044 &  0.7823      \\
\cmidrule(l){2-13} 
& \textbf{LiLT[}InfoXLM\textbf{]}$_{\mathrm{BASE}}$             &     \textbf{English only}    &   \textbf{11M}       &   \textbf{ 0.7407   }                   &       \textbf{ 0.8471  }            &     \textbf{0.8345  }               &     \textbf{  0.8335   }            &  \textbf{  0.8466   }               &       \textbf{ 0.8458       }       &     \textbf{  0.7878     }          &  \textbf{0.7643    }               &   \textbf{0.8125   } \\ \bottomrule 
\end{tabular}}
\caption{Multitask fine-tuning F1 accuracy on FUNSD and XFUND (fine-tuning on 8 languages all, testing on X). 
}
\label{tab:multitask}
\end{table*}

\section{Related Work}
During the past decade, deep learning methods became the mainstream for 
document understanding tasks \cite{yangrw4,augustorw5,siegel2018extractingrw6}.
Grid-based methods \cite{katti2018chargridrw6,denk2019bertgrid,lin2021vibertgrid} were proposed for 2D document representation where text pixels were encoded using character or word embeddings and classified into specific field types, using a convolutional neural network. GNN-based approaches \cite{gcn,yu2021pick,tang2021matchvie} adopted multi-modal features of text segments as nodes to model the document graph, and used graph neural networks to propagate information between neighboring nodes to attain a richer representation.

In recent years, self-supervised pre-training has achieved great success. Inspired by the development of the pre-trained language models in various NLP tasks, recent studies on structured document pre-training \cite{Layoutlm,layoutlmv2,xu2021layoutxlm,li2021structurallm,li2021selfdoc,li2021structext,appalaraju2021docformer}  have pushed the limits. 
LayoutLM \cite{Layoutlm} modified the BERT \cite{devlin2019bert} architecture by adding 2D spatial coordinate embeddings.
In comparison, our LiLT can be regarded as a more powerful and flexible solution for structured  document  understanding.  
LayoutLMv2 \cite{layoutlmv2} improved over LayoutLM by treating the visual features as separate tokens.
Furthermore, additional pre-training tasks were explored to improve the utilization  of unlabeled document data.  SelfDoc \cite{li2021selfdoc}  established the contextualization over a block of content, while StructuralLM \cite{li2021structurallm} proposed  cell-level 2D position embeddings and the corresponding pre-training objective. Recently, StrucTexT \cite{li2021structext} introduced a unified solution to efficiently extract semantic features from different levels and modalities to handle the entity labeling and entity linking tasks. DocFormer \cite{appalaraju2021docformer} designed a novel multi-modal self-attention layer capable of fusing textual, vision and spatial features.

Nevertheless, the aforementioned SDU approaches mainly focus on a single language - typically English, which is extremely limited with respect to multilingual application scenarios.  
To the best of our knowledge, LayoutXLM \cite{xu2021layoutxlm} was the only pre-existing multilingual SDU model, which adopted the multilingual textual model InfoXLM \cite{chi2021infoxlm} as the initialization, and adapted the LayoutLMv2 \cite{layoutlmv2}  framework to multilingual structured document pre-training. However, it required a heavy process of multilingual  data collection, cleaning and pre-training. 
On the contrary,  our LiLT can deal with the multilingual structured documents by pre-training on the monolingual  IIT-CDIP Test Collection 1.0 \cite{cdip} only.

\section{Conclusion}
In this paper, we present LiLT, a language-independent layout Transformer that can learn the layout knowledge from  monolingual structured documents and then generalize  it  to deal with multilingual ones.
Our framework successfully first decouples the text and layout information in pre-training and then re-couples them for fine-tuning.
Experimental results on eight languages under three settings (language-specific, cross-lingual zero-shot transfer, and multi-task fine-tuning) have fully illustrated 
its effectiveness,
which substantially  bridges the language gap in real-world structured document understanding  applications. 
The public availability of LiLT is also expected to promote the development of document intelligence.

For future research, we will continue to follow the pattern of transferring from ``monolingual” to ``multilingual” and 
further unlock the power of LiLT. 
In addition, we will also explore the generalized rather than language-specific visual information contained in multilingual structured documents.
\section{Acknowledgement}
This research is supported in part by NSFC (Grant No.: 61936003) and GD-NSF (No.  2017A030312006).

\bibliography{anthology,custom}

\begin{thebibliography}{45}
\expandafter\ifx\csname natexlab\endcsname\relax\def\natexlab#1{#1}\fi

\bibitem[{Afzal et~al.(2017)Afzal, K{\"o}lsch, Ahmed, and Liwicki}]{rvl-vgg}
Muhammad~Zeshan Afzal, Andreas K{\"o}lsch, Sheraz Ahmed, and Marcus Liwicki.
  2017.
\newblock Cutting the error by half: Investigation of very deep {CNN} and
  advanced training strategies for document image classification.
\newblock In \emph{ICDAR}, volume~1, pages 883--888.

\bibitem[{Appalaraju et~al.(2021)Appalaraju, Jasani, Kota, Xie, and
  Manmatha}]{appalaraju2021docformer}
Srikar Appalaraju, Bhavan Jasani, Bhargava~Urala Kota, Yusheng Xie, and
  R~Manmatha. 2021.
\newblock {DocFormer}: End-to-end {Transformer} for document understanding.
\newblock In \emph{ICCV}.

\bibitem[{Augusto Borges~Oliveira et~al.(2017)}]{augustorw5}
Dario Augusto Borges~Oliveira et~al. 2017.
\newblock {Fast CNN-based} document layout analysis.
\newblock In \emph{ICCV Workshop}, pages 1173--1180.

\bibitem[{Ba et~al.(2016)Ba, Kiros, and Hinton}]{ln}
Jimmy~Lei Ba, Jamie~Ryan Kiros, and Geoffrey~E Hinton. 2016.
\newblock Layer normalization.
\newblock \emph{arXiv preprint arXiv:1607.06450}.

\bibitem[{Bao et~al.(2020)Bao, Dong, Wei, Wang, Yang, Liu, Wang, Gao, Piao,
  Zhou et~al.}]{bao2020unilmv2}
Hangbo Bao, Li~Dong, Furu Wei, Wenhui Wang, Nan Yang, Xiaodong Liu, Yu~Wang,
  Jianfeng Gao, Songhao Piao, Ming Zhou, et~al. 2020.
\newblock {UniLMv2}: Pseudo-masked language models for unified language model
  pre-training.
\newblock In \emph{ICML}, pages 642--652.

\bibitem[{Chi et~al.(2021)Chi, Dong, Wei, Yang, Singhal, Wang, Song, Mao,
  Huang, and Zhou}]{chi2021infoxlm}
Zewen Chi, Li~Dong, Furu Wei, Nan Yang, Saksham Singhal, Wenhui Wang, Xia Song,
  Xian-Ling Mao, He-Yan Huang, and Ming Zhou. 2021.
\newblock {InfoXLM}: An information-theoretic framework for cross-lingual
  language model pre-training.
\newblock In \emph{NAACL-HLT}, pages 3576--3588.

\bibitem[{Conneau et~al.(2020)Conneau, Khandelwal, Goyal, Chaudhary, Wenzek,
  Guzm{\'a}n, Grave, Ott, Zettlemoyer, and Stoyanov}]{conneau2020unsupervised}
Alexis Conneau, Kartikay Khandelwal, Naman Goyal, Vishrav Chaudhary, Guillaume
  Wenzek, Francisco Guzm{\'a}n, {\'E}douard Grave, Myle Ott, Luke Zettlemoyer,
  and Veselin Stoyanov. 2020.
\newblock Unsupervised cross-lingual representation learning at scale.
\newblock In \emph{ACL}, pages 8440--8451.

\bibitem[{Cui et~al.(2020)Cui, Che, Liu, Qin, Wang, and
  Hu}]{cui-etal-2020-revisiting}
Yiming Cui, Wanxiang Che, Ting Liu, Bing Qin, Shijin Wang, and Guoping Hu.
  2020.
\newblock Revisiting pre-trained models for {C}hinese natural language
  processing.
\newblock In \emph{Findings of EMNLP}, pages 657--668.

\bibitem[{Das et~al.(2018)Das, Roy, Bhattacharya, and Parui}]{das2018document}
Arindam Das, Saikat Roy, Ujjwal Bhattacharya, and Swapan~K Parui. 2018.
\newblock Document image classification with intra-domain transfer learning and
  stacked generalization of deep convolutional neural networks.
\newblock In \emph{ICPR}, pages 3180--3185.

\bibitem[{Dauphinee et~al.(2019)Dauphinee, Patel, and Rashidi}]{ensemble}
Tyler Dauphinee, Nikunj Patel, and Mohammad Rashidi. 2019.
\newblock Modular multimodal architecture for document classification.
\newblock \emph{arXiv preprint arXiv:1912.04376}.

\bibitem[{Denk and Reisswig(2019)}]{denk2019bertgrid}
Timo~I Denk and Christian Reisswig. 2019.
\newblock {BERTgrid}: Contextualized embedding for {2D} document representation
  and understanding.
\newblock In \emph{Workshop on Document Intelligence at NeurIPS}.

\bibitem[{Devlin et~al.(2019)Devlin, Chang, Lee, and
  Toutanova}]{devlin2019bert}
Jacob Devlin, Ming-Wei Chang, Kenton Lee, and Kristina Toutanova. 2019.
\newblock {BERT}: Pre-training of deep bidirectional {Transformers} for
  language understanding.
\newblock In \emph{NAACL-HLT}, pages 4171--4186.

\bibitem[{Garncarek et~al.(2021)Garncarek, Powalski, Stanis{\l}awek, Topolski,
  Halama, and Grali{\'n}ski}]{garncarek2020lambert}
{\L}ukasz Garncarek, Rafa{\l} Powalski, Tomasz Stanis{\l}awek, Bartosz
  Topolski, Piotr Halama, and Filip Grali{\'n}ski. 2021.
\newblock {LAMBERT}: Layout-aware (language) modeling using {BERT} for
  information extraction.
\newblock In \emph{ICDAR}.

\bibitem[{Harley et~al.(2015)}]{rvl}
Adam~W Harley et~al. 2015.
\newblock Evaluation of deep convolutional nets for document image
  classification and retrieval.
\newblock In \emph{ICDAR}, pages 991--995.

\bibitem[{Hong et~al.(2020)Hong, Kim, Ji, Hwang, Nam, and Park}]{hong2020bros}
Teakgyu Hong, DongHyun Kim, Mingi Ji, Wonseok Hwang, Daehyun Nam, and Sungrae
  Park. 2020.
\newblock {BROS}: A pre-trained language model for understanding texts in
  document.

\bibitem[{Jaume et~al.(2019)}]{funsd}
Guillaume Jaume et~al. 2019.
\newblock {FUNSD}: A dataset for form understanding in noisy scanned documents.
\newblock In \emph{ICDAR}, volume~2, pages 1--6.

\bibitem[{Katti et~al.(2018)Katti, Reisswig, Guder, Brarda, Bickel, H{\"o}hne,
  and Faddoul}]{katti2018chargridrw6}
Anoop~R Katti, Christian Reisswig, Cordula Guder, Sebastian Brarda, Steffen
  Bickel, Johannes H{\"o}hne, and Jean~Baptiste Faddoul. 2018.
\newblock Chargrid: Towards understanding {2D} documents.
\newblock In \emph{EMNLP}, pages 4459--4469.

\bibitem[{Kingma and Ba(2015)}]{adam}
Diederik~P Kingma and Jimmy Ba. 2015.
\newblock Adam: A method for stochastic optimization.
\newblock In \emph{ICLR}.

\bibitem[{Lample et~al.(2016)Lample, Ballesteros, Subramanian, Kawakami, and
  Dyer}]{blstmcrf}
Guillaume Lample, Miguel Ballesteros, Sandeep Subramanian, Kazuya Kawakami, and
  Chris Dyer. 2016.
\newblock Neural architectures for named entity recognition.
\newblock In \emph{NAACL-HLT}, pages 260--270.

\bibitem[{Lewis et~al.(2006)Lewis, Agam, Argamon, Frieder, Grossman, and
  Heard}]{cdip}
David Lewis, Gady Agam, Shlomo Argamon, Ophir Frieder, David Grossman, and
  Jefferson Heard. 2006.
\newblock Building a test collection for complex document information
  processing.
\newblock In \emph{ACM SIGIR}, pages 665--666.

\bibitem[{Li et~al.(2021{\natexlab{a}})Li, Bi, Yan, Wang, Huang, Huang, and
  Si}]{li2021structurallm}
Chenliang Li, Bin Bi, Ming Yan, Wei Wang, Songfang Huang, Fei Huang, and Luo
  Si. 2021{\natexlab{a}}.
\newblock {StructuralLM}: Structural pre-training for form understanding.
\newblock In \emph{ACL}.

\bibitem[{Li et~al.(2021{\natexlab{b}})Li, Gu, Kuen, Morariu, Zhao, Jain,
  Manjunatha, and Liu}]{li2021selfdoc}
Peizhao Li, Jiuxiang Gu, Jason Kuen, Vlad~I Morariu, Handong Zhao, Rajiv Jain,
  Varun Manjunatha, and Hongfu Liu. 2021{\natexlab{b}}.
\newblock {SelfDoc}: Self-supervised document representation learning.
\newblock In \emph{CVPR}, pages 5652--5660.

\bibitem[{Li et~al.(2021{\natexlab{c}})Li, Qian, Yu, Qin, Zhang, Liu, Yao, Han,
  Liu, and Ding}]{li2021structext}
Yulin Li, Yuxi Qian, Yuchen Yu, Xiameng Qin, Chengquan Zhang, Yan Liu, Kun Yao,
  Junyu Han, Jingtuo Liu, and Errui Ding. 2021{\natexlab{c}}.
\newblock {StrucTexT}: Structured text understanding with multi-modal
  {Transformers}.
\newblock In \emph{ACM-MM}.

\bibitem[{Lin et~al.(2017)Lin, Doll{\'a}r, Girshick, He, Hariharan, and
  Belongie}]{fpn}
Tsung-Yi Lin, Piotr Doll{\'a}r, Ross Girshick, Kaiming He, Bharath Hariharan,
  and Serge Belongie. 2017.
\newblock Feature pyramid networks for object detection.
\newblock In \emph{CVPR}, pages 2117--2125.

\bibitem[{Lin et~al.(2021)Lin, Gao, Sun, Zhong, Hu, Ren, and
  Huo}]{lin2021vibertgrid}
Weihong Lin, Qifang Gao, Lei Sun, Zhuoyao Zhong, Kai Hu, Qin Ren, and Qiang
  Huo. 2021.
\newblock {ViBERTgrid}: A jointly trained multi-modal {2D} document
  representation for key information extraction from documents.
\newblock In \emph{ICDAR}.

\bibitem[{Liu et~al.(2019{\natexlab{a}})Liu, Gao, Zhang, and Zhao}]{gcn}
Xiaojing Liu, Feiyu Gao, Qiong Zhang, and Huasha Zhao. 2019{\natexlab{a}}.
\newblock Graph convolution for multimodal information extraction from visually
  rich documents.
\newblock In \emph{NAACL-HLT}, pages 32--39.

\bibitem[{Liu et~al.(2019{\natexlab{b}})Liu, Ott, Goyal, Du, Joshi, Chen, Levy,
  Lewis, Zettlemoyer, and Stoyanov}]{liu2019roberta}
Yinhan Liu, Myle Ott, Naman Goyal, Jingfei Du, Mandar Joshi, Danqi Chen, Omer
  Levy, Mike Lewis, Luke Zettlemoyer, and Veselin Stoyanov. 2019{\natexlab{b}}.
\newblock {RoBERTa}: A robustly optimized {BERT} pretraining approach.
\newblock \emph{arXiv preprint arXiv:1907.11692}.

\bibitem[{Loshchilov and Hutter(2018)}]{adamw}
Ilya Loshchilov and Frank Hutter. 2018.
\newblock Decoupled weight decay regularization.
\newblock In \emph{ICLR}.

\bibitem[{Lu et~al.(2019)Lu, Batra, Parikh, and Lee}]{lu2019vilbert}
Jiasen Lu, Dhruv Batra, Devi Parikh, and Stefan Lee. 2019.
\newblock {ViLBERT}: Pretraining task-agnostic visiolinguistic representations
  for vision-and-language tasks.
\newblock \emph{NeurIPS}, 32:13--23.

\bibitem[{Park et~al.(2019)Park, Shin, Lee, Lee, Surh, Seo, and
  Lee}]{park2019cord}
Seunghyun Park, Seung Shin, Bado Lee, Junyeop Lee, Jaeheung Surh, Minjoon Seo,
  and Hwalsuk Lee. 2019.
\newblock {CORD}: A consolidated receipt dataset for post-{OCR} parsing.
\newblock In \emph{Workshop on Document Intelligence at NeurIPS}.

\bibitem[{Powalski et~al.(2021)Powalski, Borchmann, Jurkiewicz, Dwojak,
  Pietruszka, and Pa{\l}ka}]{2021going}
Rafa{\l} Powalski, {\L}ukasz Borchmann, Dawid Jurkiewicz, Tomasz Dwojak,
  Micha{\l} Pietruszka, and Gabriela Pa{\l}ka. 2021.
\newblock Going full-{TILT} boogie on document understanding with
  text-image-layout {Transformer}.
\newblock In \emph{ICDAR}.

\bibitem[{Qian et~al.(2019)Qian, Santus, Jin, Guo, and
  Barzilay}]{qian2019graphie}
Yujie Qian, Enrico Santus, Zhijing Jin, Jiang Guo, and Regina Barzilay. 2019.
\newblock {GraphIE}: A graph-based framework for information extraction.
\newblock In \emph{NAACL-HLT}, pages 751--761.

\bibitem[{Sarkhel and Nandi(2019)}]{ladder}
Ritesh Sarkhel and Arnab Nandi. 2019.
\newblock Deterministic routing between layout abstractions for multi-scale
  classification of visually rich documents.
\newblock In \emph{IJCAI}, pages 3360--3366.

\bibitem[{Siegel et~al.(2018)Siegel, Lourie, Power, and
  Ammar}]{siegel2018extractingrw6}
Noah Siegel, Nicholas Lourie, Russell Power, and Waleed Ammar. 2018.
\newblock Extracting scientific figures with distantly supervised neural
  networks.
\newblock In \emph{JCDL}, pages 223--232.

\bibitem[{Szegedy et~al.(2017)Szegedy, Ioffe, Vanhoucke, and
  Alemi}]{szegedy2017inception}
Christian Szegedy, Sergey Ioffe, Vincent Vanhoucke, and Alexander~A Alemi.
  2017.
\newblock {Inception-v4, Inception-ResNet} and the impact of residual
  connections on learning.
\newblock In \emph{AAAI}, pages 4278--4284.

\bibitem[{Tang et~al.(2021)Tang, Xie, Jin, Wang, Chen, Xu, Wang, Wu, and
  Li}]{tang2021matchvie}
Guozhi Tang, Lele Xie, Lianwen Jin, Jiapeng Wang, Jingdong Chen, Zhen Xu,
  Qianying Wang, Yaqiang Wu, and Hui Li. 2021.
\newblock {MatchVIE}: Exploiting match relevancy between entities for visual
  information extraction.
\newblock In \emph{IJCAI}, pages 1039--1045.

\bibitem[{Wang et~al.(2021{\natexlab{a}})Wang, Liu, Jin, Tang, Zhang, Zhang,
  Wang, Wu, and Cai}]{vies}
Jiapeng Wang, Chongyu Liu, Lianwen Jin, Guozhi Tang, Jiaxin Zhang, Shuaitao
  Zhang, Qianying Wang, Yaqiang Wu, and Mingxiang Cai. 2021{\natexlab{a}}.
\newblock Towards robust visual information extraction in real world: New
  dataset and novel solution.
\newblock In \emph{AAAI}, volume~35, pages 2738--2745.

\bibitem[{Wang et~al.(2021{\natexlab{b}})Wang, Wang, Tang, Jin, Ma, Ding, and
  Huang}]{tcpn}
Jiapeng Wang, Tianwei Wang, Guozhi Tang, Lianwen Jin, Weihong Ma, Kai Ding, and
  Yichao Huang. 2021{\natexlab{b}}.
\newblock Tag, copy or predict: A unified weakly-supervised learning framework
  for visual information extraction using sequences.
\newblock In \emph{IJCAI}, pages 1082--1090.

\bibitem[{Xie et~al.(2017)Xie, Girshick, Doll{\'a}r, Tu, and He}]{resnext}
Saining Xie, Ross Girshick, Piotr Doll{\'a}r, Zhuowen Tu, and Kaiming He. 2017.
\newblock Aggregated residual transformations for deep neural networks.
\newblock In \emph{CVPR}, pages 1492--1500.

\bibitem[{Xu et~al.(2021{\natexlab{a}})Xu, Xu, Lv, Cui, Wei, Wang, Lu,
  Florencio, Zhang, Che et~al.}]{layoutlmv2}
Yang Xu, Yiheng Xu, Tengchao Lv, Lei Cui, Furu Wei, Guoxin Wang, Yijuan Lu,
  Dinei Florencio, Cha Zhang, Wanxiang Che, et~al. 2021{\natexlab{a}}.
\newblock {LayoutLMv2}: Multi-modal pre-training for visually-rich document
  understanding.
\newblock In \emph{ACL}.

\bibitem[{Xu et~al.(2020)Xu, Li, Cui, Huang, Wei, and Zhou}]{Layoutlm}
Yiheng Xu, Minghao Li, Lei Cui, Shaohan Huang, Furu Wei, and Ming Zhou. 2020.
\newblock {LayoutLM}: Pre-training of text and layout for document image
  understanding.
\newblock In \emph{ACM-SIGKDD}, pages 1192--1200.

\bibitem[{Xu et~al.(2021{\natexlab{b}})Xu, Lv, Cui, Wang, Lu, Florencio, Zhang,
  and Wei}]{xu2021layoutxlm}
Yiheng Xu, Tengchao Lv, Lei Cui, Guoxin Wang, Yijuan Lu, Dinei Florencio, Cha
  Zhang, and Furu Wei. 2021{\natexlab{b}}.
\newblock {LayoutXLM}: Multimodal pre-training for multilingual visually-rich
  document understanding.
\newblock \emph{arXiv preprint arXiv:2104.08836}.

\bibitem[{Yang et~al.(2017)Yang, Yumer, Asente, Kraley, Kifer, and
  Lee~Giles}]{yangrw4}
Xiao Yang, Ersin Yumer, Paul Asente, Mike Kraley, Daniel Kifer, and
  C~Lee~Giles. 2017.
\newblock Learning to extract semantic structure from documents using
  multimodal fully convolutional neural networks.
\newblock In \emph{CVPR}, pages 5315--5324.

\bibitem[{Yu et~al.(2021)Yu, Lu, Qi, Gong, and Xiao}]{yu2021pick}
Wenwen Yu, Ning Lu, Xianbiao Qi, Ping Gong, and Rong Xiao. 2021.
\newblock {PICK}: Processing key information extraction from documents using
  improved graph learning-convolutional networks.
\newblock In \emph{ICPR}, pages 4363--4370.

\bibitem[{Zhang et~al.(2020)Zhang, Xu, Cheng, Pu, Lu, Qiao, Niu, and Wu}]{TRIE}
Peng Zhang, Yunlu Xu, Zhanzhan Cheng, Shiliang Pu, Jing Lu, Liang Qiao, Yi~Niu,
  and Fei Wu. 2020.
\newblock {TRIE}: End-to-end text reading and information extraction for
  document understanding.
\newblock In \emph{ACM-MM}, pages 1413--1422.

\end{thebibliography}
\bibliographystyle{acl_natbib}

\clearpage
\appendix
\setcounter{secnumdepth}{0}
\section{Appendix}
\setcounter{secnumdepth}{1}
\section{Dataset Details}\label{app1}

\paragraph{FUNSD} FUNSD \cite{funsd} is an English dataset for form understanding in noisy scanned documents. It contains 199 real, fully annotated, scanned forms where 9,707 semantic entities are annotated above 31,485 words. The 199 samples are split into 149 for training and 50 for testing.
We directly use the official OCR annotations.
The semantic entity recognition (SER) task is assigning to each word a semantic entity label from a set of four predefined categories: question, answer, header, or other. The entity-level F1 score is used as the evaluation metric (Table \ref{tab:funsd}).

\paragraph{CORD} CORD \cite{park2019cord} is an English receipt dataset for key information extraction. Its publicly available subset  includes 800 receipts for the training set, 100 for the validation set, and 100 for the test set. A photo and a list of OCR annotations are equipped for each receipt. 
The dataset defines 30 fields under 4 categories and the task aims to label each word to the right field. The evaluation metric is the entity-level F1 score, as shown in Table \ref{tab:cord}. We use the official OCR annotations.

\paragraph{EPHOIE} EPHOIE \cite{vies} is collected from actual Chinese examination papers with the diversity of text types and layout distribution. The 1,494 samples are divided into a training set with 1,183 images and a testing set with 311 images, respectively. 
It defines ten entity categories, and we provide the entity-level F1  score for RoBERTa, LayoutXLM and LiLT in Table \ref{tab:ephoie}.  The official OCR annotations are adopted.

\paragraph{RVL-CDIP} RVL-CDIP \cite{rvl} consists of 400,000 gray-scale images of English documents, with 8:1:1 for the training set, validation set, and test set. A multi-class single-label classification task is defined on RVL-CDIP. The images are categorized into 16 classes, with 25,000 images per class. The evaluation metric is the overall classification accuracy as shown in Table \ref{tab:rvl}. Text and layout information are extracted by TextIn API.

\paragraph{XFUND} XFUND \cite{xu2021layoutxlm} is a multilingual form understanding dataset that contains 1,393 fully annotated forms with seven languages including Chinese (ZH), Japanese (JA), Spanish (ES), French (FR), Italian (IT), German (DE), and Portuguese (PT). Each language
includes 199 forms, where the training set includes
149 forms, and the test set includes 50 forms. We focus on the semantic entity recognition (SER) and relation extraction (RE) tasks defined in the original paper \cite{xu2021layoutxlm}. Relation extraction aims to predict the relation between any two given semantic entities, and we mainly focus on the key-value relation extraction. We use the official OCR results, and the same F1 accuracy  evaluation metric as in LayoutXLM  \cite{xu2021layoutxlm} for Table \ref{tab:language-specific}, \ref{tab:zero-shot} and \ref{tab:multitask}.

\section{Fine-tuning Details}\label{app2}
\paragraph{Fine-tuning for Semantic Entity Recognition} We conduct the semantic entity recognition task on FUNSD, CORD, EPHOIE and XFUND. 
We build a token-level classification layer above the output representations to predict the BIO tags for each entity field.

\paragraph{Fine-tuning for Document Classification}
This task depends on high-level visual information, thereby we leverage the image features explicitly in the fine-tuning stage, following LayoutLMv2 \cite{layoutlmv2}. We pool the visual feature of the ResNeXt101-FPN \cite{resnext,fpn} backbone into a global  feature,
concatenate it with the \texttt{[CLS]} output feature,  and feed them into the final classification layer. 

\paragraph{Fine-tuning for Relation Extraction}
We build the additional  head for relation extraction on the FUNSD and XFUND datasets following \cite{xu2021layoutxlm} for fair comparison. We first incrementally construct the set of relation candidates by producing all possible pairs of given semantic entities. For every pair, the representation of the head/tail entity is the concatenation of the first token vector in each entity and the entity type embedding obtained with a specific type embedding layer. After respectively projected by two FFN layers, the representations of head and tail are concatenated and then fed into a bi-affine classifier.

\end{document}